\documentclass[10pt,twocolumn,letterpaper]{article}

\usepackage[pagenumbers]{cvpr} 

\definecolor{cvprblue}{rgb}{0.21,0.49,0.74}
\usepackage[pagebackref,breaklinks,colorlinks,allcolors=cvprblue]{hyperref}
\usepackage{bm}
\usepackage{multirow}
\newcommand{\ourmodel}{X-Dyna\xspace}
\newcommand{\mixdatatraining}{Harmonic Data Fusion Training\xspace}
\newcommand{\xbody}{S-Face ControlNet\xspace}
\newcommand{\dynamicsadapter}{Dynamics-Adapter\xspace}
\newcommand{\faceencoder}{Face-ID-Adapter\xspace}

\def\paperID{134} 
\def\confName{CVPR}
\def\confYear{2025}
\def\papername{X-Dyna}

\title {\textcolor{Red}{X}-\textcolor{Orange}{Dyna}: \textcolor{Red}{Ex}pressive \textcolor{Orange}{Dyna}mic Human Image Animation}

\author{Di Chang$^{1,2}$ \hspace{12pt} Hongyi Xu$^{2*}$\hspace{12pt} You Xie$^{2*}$ \hspace{12pt}Yipeng Gao$^{1*}$ \hspace{12pt}Zhengfei Kuang$^{3*}$ \hspace{12pt}Shengqu Cai$^{3*}$\\
 Chenxu Zhang$^{2*}$\hspace{12pt} Guoxian Song$^{2}$\hspace{12pt} Chao Wang$^{2}$ \hspace{12pt} Yichun Shi$^{2}$ \hspace{12pt} Zeyuan Chen$^{2,5}$\\
 Shijie Zhou$^{4}$ \hspace{12pt} Linjie Luo$^{2}$ \hspace{12pt}  Gordon Wetzstein$^{3}$  \hspace{12pt} Mohammad Soleymani$^{1}$ \\
\small{$^1$ University of Southern California\hspace{13pt} $^2$ ByteDance \hspace{13pt} $^3$ Stanford University}\\
 \small{$^4$ University of California Los Angeles\hspace{9pt} $^5$ University of California San Diego} \\
\url{https://x-dyna.github.io/xdyna.github.io/} \\
\tt\small dichang@usc.edu\\}

\begin{document}
\maketitle

\def\thefootnote{*}\footnotetext{Equally contributed as second authors}

\begin{abstract} 
\label{abstract}

We introduce \papername, a novel zero-shot, diffusion-based pipeline for animating a single human image using facial expressions and body movements derived from a driving video, 
that generates realistic, context-aware dynamics for both the subject and the surrounding environment. 
Building on prior approaches centered on human pose control, \papername~addresses key shortcomings causing the loss of dynamic details, enhancing the lifelike qualities of human video animations.
At the core of our approach is the \textbf{\dynamicsadapter}, a lightweight module that effectively integrates reference appearance context into the spatial attentions of the diffusion backbone while preserving the capacity of motion modules in synthesizing fluid and intricate dynamic details. Beyond body pose control, we connect a local control module with our model to capture identity-disentangled facial expressions, facilitating accurate expression transfer for enhanced realism in animated scenes. 
Together, these components form a unified framework capable of learning physical human motion and natural scene dynamics from a diverse blend of human and scene videos.
Comprehensive qualitative and quantitative evaluations demonstrate that \ourmodel outperforms state-of-the-art methods, creating highly lifelike and expressive animations. The code is available at \url{https://github.com/bytedance/X-Dyna}.

\end{abstract}    
\section{Introduction}
\label{intro}
\begin{figure*}[t!]\vspace{-5pt}
\centering
 \includegraphics[width=\linewidth]{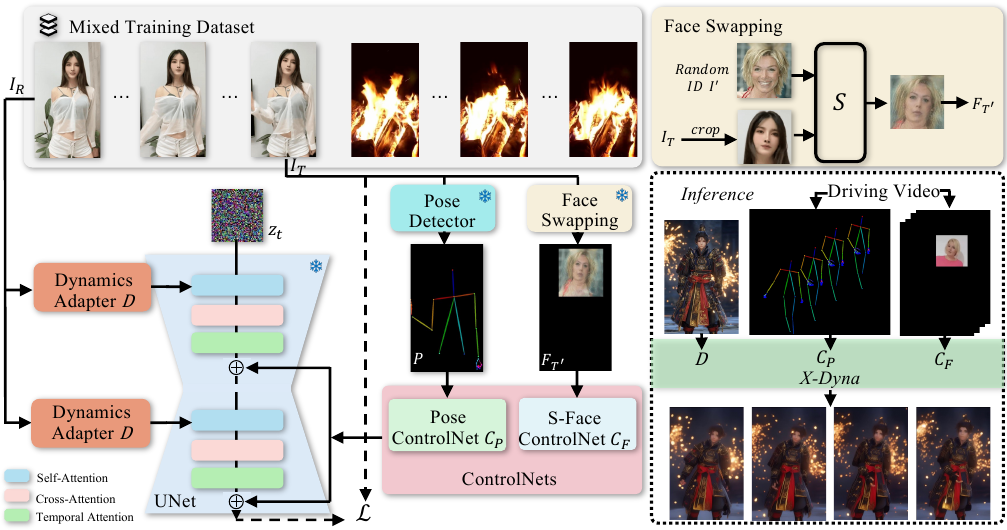}
   \caption{ 
   We leverage a pretrained diffusion UNet backbone for controlled human image animation, enabling expressive dynamic details and precise motion control. Specifically, we introduce a dynamics adapter $D$ that seamlessly integrates the reference image context as a trainable residual to the spatial attention, in parallel with the denoising process, while preserving the original spatial and temporal attention mechanisms within the UNet. In addition to body pose control via a ControlNet $C_P$ , we introduce a local face control module $C_F$ that implicitly learns facial expression control from a synthesized cross-identity face patch. We train our model on a diverse dataset of human motion videos and natural scene videos simultaneously. Our model achieves remarkable transfer of body poses and facial expressions, as well as highly vivid and detailed dynamics for both the human and the scene.
   }
    \label{fig:pipeline}
\end{figure*}





We investigate the task of human video generation, focusing on animating a single human image using body movements and facial expressions derived from a driving video of a different person. This area has garnered growing interest owing to its numerous applications in digital arts, social media and virtual humans. Building on prior research~\cite{hu2024animate,xu2024magicanimate,chang2023magicpose,zhang2024mimicmotion,zhu2024champ,wang2024vividpose,musepose}, our goal is to advance the field of zero-shot human image animation by not only enhancing the accuracy of pose and expression transfer but also by incorporating vivid human dynamics, e.g., blowing hair and flowing garments, and natural environmental effects, e.g., waterfalls, rain, and fireworks.


Recent approaches have tackled human image animation as a controlled image-to-video diffusion task. 
These methods typically employ a parallel UNet to incorporate reference appearance through mutual self-attentions~\cite{cao2023masactrl}, while body motion cues (e.g., 2D skeletons~\cite{chang2023magicpose,musepose,zhang2024mimicmotion} and DensePose~\cite{xu2024magicanimate}) are integrated as spatial guidance through frameworks like ControlNet~\cite{zhang2023adding} or PoseGuider~\cite{hu2024animate}. Temporal modules, such as AnimateDiff~\cite{guo2023animatediff} and Align-Your-Latents~\cite{blattmann2023align,blattmann2023stable}, have been introduced to the diffusion backbone, trained from large-scale videos to enhance consistency and dynamics in visual sequence generation. 
Despite improvements in control precision and generation realism,  the combined modules for human image animation often fall short in capturing intricate visual dynamics, leading to static backgrounds and rigid human motions. This shortcoming, rooted in both network design and training data distribution, ultimately compromises the lifelike quality of the generated videos.

To this end, we propose~\textbf{\papername}, a diffusion-based human image animation pipeline that achieves the accurate transfer of pose and facial expressions along with consistent and vivid human and background dynamics. We observe that the loss of dynamic details  primarily arises from the strong appearance constraints on spatial attentions imposed by the appearance reference modules, typically formulated as a trainable copy of a parallel UNet. To address this, we introduce a lightweight cross-frame attention module, \textit{Dynamics-Adapter}, which seamlessly propagates the reference appearance context to the denoising process by feeding the denoised reference image in parallel with noised sequences to the model. It integrates with the diffusion backbone via a trainable copy of the query projector and zero-initialized output projector, ensuring the backbone’s spatial and temporal generation capability stays intact. Unlike the standard I2V settings~\cite{blattmann2023stable} that generate subsequent frames from the reference image, our design maintains appearance consistency from varying poses, in coordination with pose control modules. Notably, beyond body pose control, we employ a local control module to capture identity-disentangled facial expressions, enhancing realism with accurate expression transfer. While prior image animation models are primarily trained on human videos with static backgrounds, our dynamics adapter enables the learning of subtle human dynamics and fluid environmental effects, in addition to body and facial expression controls,  from a diverse mixture of human and scene videos.

Trained on a curated dataset of 900h human dancing and natural scene videos, our method excels at accurately transferring the body poses and facial expressions while generating lifelike human and scene dynamics consistent with the reference image context. We comprehensively evaluate our model on challenging benchmarks~\cite{Jafarian_2021_CVPR_TikTok,pexels,midjourney}, and~\papername~outperforms state-of-the-art human image animation baselines both quantitatively and qualitatively, demonstrating superior dynamics expressiveness, identity preservation and visual quality. Our main contributions are: 
\begin{itemize}[nolistsep,leftmargin=*]
\item A zero-shot diffusion-based human image animation model for both pose control and dynamics synthesis, trained on a mixture of human and natural scene videos;

\item An efficient Dynamics-Adapter module that effectively incorporates reference appearance while maintaining the foundation model’s capability in generating high-quality dynamics;

\item A local implicit face control module that enables refined, identity-disentangled facial expression control; and

\item Demonstration of captivating zero-shot controllable human image animations and live photos with vivid dynamics. 

\end{itemize}

\begin{figure*}[t!]\vspace{-5pt}
\centering
 \includegraphics[width=0.98\linewidth]{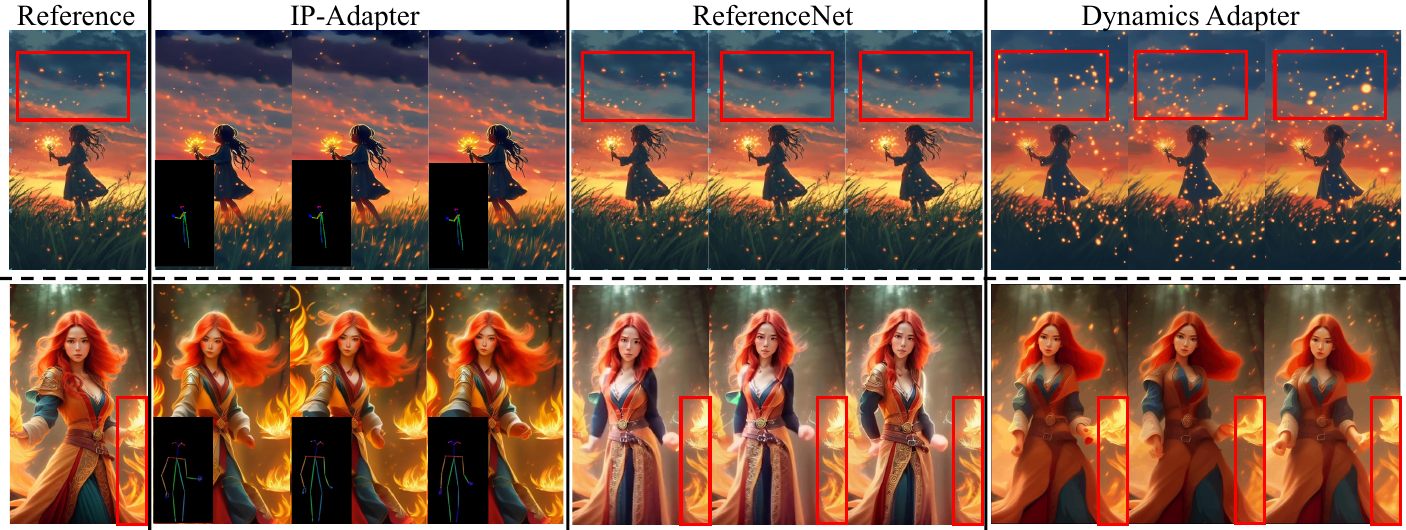}
  \vspace{-5pt}
   \caption{ a) IP-Adapter~\cite{ye2023ip} can generate vivid texture from the reference image but fails to preserve the appearance. b) Though ReferenceNet~\cite{hu2024animate} can preserve the identity from the human reference, it generates a static background without any dynamics. c) \dynamicsadapter provides both expressive details and consistent identities.
   }
    \label{fig:architecture_comparison_vis}
\end{figure*}

\begin{figure}[t!]\vspace{-5pt}
\centering
 \includegraphics[width=\linewidth]{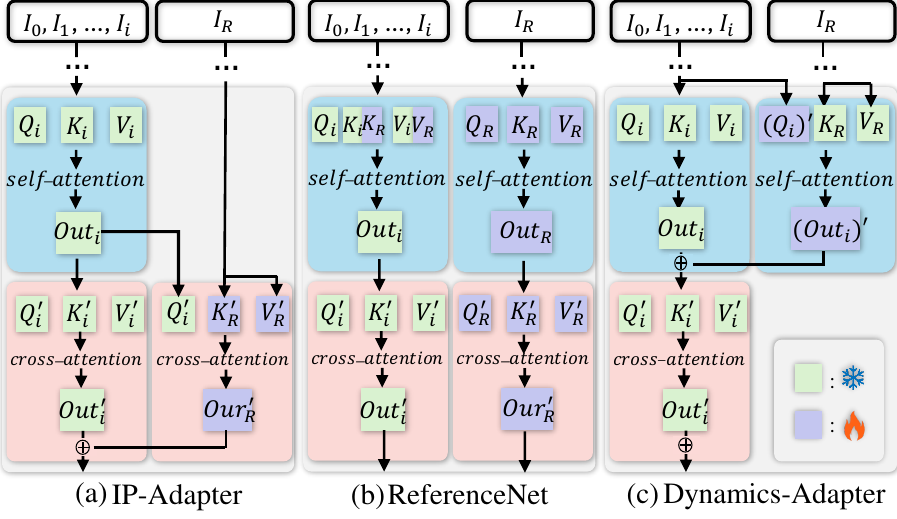}
  \vspace{-10pt}
   \caption{a) IP-Adapter~\cite{ye2023ip} encodes the reference image as an image CLIP embedding and injects the information into the cross-attention layers in SD as the residual. b) ReferenceNet~\cite{hu2024animate} is a trainable parallel UNet and feeds the semantic information into SD via concatenation of self-attention features. c) \dynamicsadapter encodes the reference image with a partially shared-weight UNet. The appearance control is realized by learning a residual in the self-attention with trainable query and output linear layers. All other components share the same frozen weight with SD.}
   \vspace{-10pt}
    \label{fig:architecture_comparison}
\end{figure}

\section{Related Works}
\label{related}
\subsection{Diffusion Models for Human Video Animation}
Recent advancements~\cite{po2024state} in latent diffusion models~\cite{rombach2022high} have greatly advanced human image animation. Previous approaches~\cite{wang2023disco,chang2023magicpose,hu2024animate,xu2024magicanimate} commonly employed a two-stage training paradigm: in the first stage, a pose-driven image model is trained on individual video frames paired with corresponding pose images; in the second stage, a temporal module is introduced to capture temporal dynamics, while the image generation model remains fixed. Following this framework, these methods integrate ReferenceNet~\cite{hu2024animate} with a UNet architecture to extract appearance features from reference characters. With progress in the video foundation models, recent works~\cite{wang2024vividpose,zhang2024mimicmotion} have simplified this process by directly fine-tuning Stable Video Diffusion~\cite{blattmann2023stable}, effectively replacing the two-stage training approach.
As mentioned in Sec.~\ref{intro}, there are several human video animation methods~\cite{hu2024animate,xu2024magicanimate,chang2023magicpose,wang2023disco,karras2023dreampose,liu2024disentangling,ma2024follow,wang2024vividpose,zhang2024mimicmotion,zhu2024champ,musepose,musev}, including CLIP~\cite{radford2021learning} embedding with ControlNet~\cite{zhang2023adding}, ReferenceNet~\cite{hu2024animate,chang2023magicpose,xu2024magicanimate,zhu2024champ} with ControlNet~\cite{zhang2023adding}, and SVD~\cite{blattmann2023stable} with Pose Encoder~\cite{wei2024aniportrait}. However, these methods are not capable of capturing dynamics-related semantic information from the reference image and cannot provide a vivid animation of physical details from a natural background and human foreground.
\subsection{Dynamics Generation}
Dynamics generation has become a critical area in video generation, focusing on creating realistic motion and temporal consistency. GAN-based methods such as TGAN~\cite{saito2017temporal} and MoCoGAN~\cite{tulyakov2018mocogan} pioneered the decomposition of motion and content, allowing for better temporal coherence. However, GANs often struggle with complex motion scenes, and artifacts may appear due to difficulties in modeling long-term dependencies. Later models, such as Progressive Growing of GANs~\cite{karras2017progressive}, introduced gradual increases in resolution, achieving more stable results in video synthesis.
Diffusion models have emerged as powerful alternatives for video generation, with methods like Video Diffusion Models~\cite{ho2022video} and AnimateDiff~\cite{guo2023animatediff} incorporating temporal conditioning to ensure consistency across frames. AnimateDiff, for instance, applies temporal attention to produce smoother, continuous animations in human-centered videos. Similarly, Stable Video Diffusion~\cite{blattmann2023stable} employs temporal modeling strategies to enhance dynamic texture quality, often surpassing GAN-based approaches in long-term coherence and photorealism. These works inspired most recent diffusion-based methods~\cite{li2024generative,feng2024explorative} to further improve the ability for dynamics generation.

\section{Method}
\label{method}
Given a single reference image $I_{R}$, the objective of \papername~is to reanimate the human subject with a pose and expression sequence ${P_i}$ derived from a driving video, where $i = 1, \ldots, T$ denotes the frame index. Most prior approaches decompose this task into two main sub-tasks: (1) transferring the appearance of the individual and background from the reference image and (2) controlling the video frames based on the pose and expression sequence ${P_i}$.  \papername~not only focuses on generating temporally smooth image sequences but also aims to enhance lifelike dynamics realism. We achieve this by creating vivid and expressive dynamics for both the foreground human and background scenes in an end-to-end fashion, eliminating the need for any foreground and background disentanglement pre- or post-processing steps.

In Section.~\ref{architectures}, we first examine existing network designs for transferring reference appearance and background, and identity their underlying causes for the loss of dynamic details. We then introduce our dynamics-adapter, which achieves accurate transfer of reference appearance with minimal impact on the diffusion backbone’s dynamics synthesis capability. To further enhance expression transfer and identity preservation, we integrate an additional local control module using synthetic cross-driven face images, as elaborated in Section~\ref{face}. Our model design enables us to effectively learn human dynamics and environmental effects simultaneously from a diverse fusion of human motion and natural scenes videos (Section~\ref{fusion}).  Our pipeline is illustrated in Figure.~\ref{fig:pipeline}. 


\subsection{Preliminary}

\paragraph{Latent Diffusion Model.} Facilitated by a pretrained auto-encoder, latent diffusion models~\cite{rombach2022high} are a class of diffusion models~\cite{ho2020denoising,song2020score,song2020denoising} that synthesize desired samples in the image latent space, starting from Gaussian noise $z_T \sim \mathcal{N}(0,1)$ and refining through $T$ denoising steps. During training, latent representations of images are progressively corrupted by Gaussian noise $\epsilon$, following the Denoising Diffusion Probabilistic Model (DDPM) framework~\cite{ho2020denoising}. A UNet-based denoising backbone network containing intervened layers of convolutions and attentions, is trained to learn the reverse denoising process. 

For our task, we employ a pretrained text-to-image (T2I) diffusion model Stable Diffusion (SD) as the generative backbone, with the addition of a ControlNet~\cite{zhang2023adding} module to incorporate 2D skeletal pose control as in recent human image animation work~\cite{chang2023magicpose,musepose,hu2024animate}, as well as temporal modules~\cite{guo2023animatediff} for enhanced consistency across generated video frames.

\paragraph{Appearance Reference.}\label{architectures}
Previous research has introduced various strategies to maintain appearance consistency with a given reference image. Early approaches such as ~\cite{wang2023disco} represent reference appearance features using CLIP image embeddings, which are injected into the text-conditioned cross-attention layers of the diffusion backbone. More recently, IP-Adapter~\cite{ye2023ip} introduced a novel approach where image CLIP embeddings are incorporated into the diffusion model via new cross-attention layers, which learn to predict a residual over the original cross-attention latents, as illustrated in (Figure~\ref{fig:architecture_comparison} (a)). However, due to limitations in the CLIP image embeddings’ ability to capture detailed appearance information, this approach often results in noticeable identity loss and inconsistencies.
The latest human image animation models~\cite{hu2024animate,chang2023magicpose,xu2024magicanimate,zhu2024champ} have addressed these shortcomings by employing a ReferenceNet module for appearance control. ReferenceNet, a parallel and trainable duplication of the entire diffusion UNet, captures rich, detailed appearance features from a single reference image and interconnects with the diffusion UNet’s self-attention layers through feature concatenation (Fig.~\ref{fig:architecture_comparison} (b)). Although this method effectively transfers appearance features to the denoising process, the full set of trainable parameters in ReferenceNet often imposes a strong and strict influence over all spatial pixels, resulting in static backgrounds and rigid dynamics, as visualized in Fig.~\ref{fig:architecture_comparison_vis}.

\subsection{Dynamics Adapter}
\label{animation}
To address the aforementioned limitations in existing reference appearance control designs, we introduce a \textit{\dynamicsadapter} module which effectively transfers human appearance and background context from the reference image to the diffusion backbone, without compromising its generative capability for dynamic motion synthesis. Inspired by the attention mechanism in the I2V-Adapter~\cite{guo2023i2v} which generates subsequent video frames from the given reference image guided by text prompts, our dynamics adapter  $\mathcal{D}$  is tailored for explicit cross-driven pose and expression control. Unlike I2V, our task accommodates motions that may differ significantly from the pose and expression of the reference image, often originating from subjects with distinct identities and body characteristics. To achieve this,  $\mathcal{D}$  is designed as a shared-weight, parallel UNet branch that injects layer-by-layer self-attention guidance of reference appearance features. 

The self-attention calculation in the transformer blocks of the diffusion UNet can be represented as:

\begin{equation}
\label{eq:sd_self_attn}
\bm{A}_i = \texttt{softmax}(\frac{\bm{Q}_{i} \bm{K}_{i}^\top}{\sqrt{d}}) \bm{V}_{i} , 
\end{equation}
where $\bm{Q}_i, \bm{K}_i, \bm{V}_i$ are query, key, and value of the $i$th latent noise frame, respectively, and $d$ is the dimension of the key and query.  To introduce reference appearance guidance through our dynamics adapter, we capitalize on the prior capabilities of the original UNet to generate the key $\bm{K}_{R}$ and value $\bm{V}_{R}$ from the denoised latent map of the reference image $\bm{I}_R. $ Additionally a trainable copy of query projector forms new query matrices $\bm{Q}^{'}_{i}$ from the latent noise of generation frame $\bm{I}_{i}.$ This enables a cross-frame attention, computed as:
\begin{equation}\label{eq:our_attn}
   \bm{A}'_i = \texttt{softmax}(\frac{\bm{Q}'_{i} \bm{K}_{R}^\top}{\sqrt{d}}) \bm{V}_{R}.
\end{equation}
We combine these two attention outputs with separate output projection matrices, $\bm{W}_{O}$ and $\bm{W}'_{O},$ as follows:

\begin{equation}
\label{eq:self_attn_out}
   \bm{Out}_i = (\bm{A}_i\bm{W}_{O}) \ +  \ (\bm{A}'_i  {\bm{W}'_{O}} ),
\end{equation}
where  $\mathbf{W}'_{O}$ is a trainable output projector. 

This residual term enriches the original spatial attentions with correlated and detailed appearance information derived from the reference image. To implement this seamlessly, we initialize the query projector weights from the original UNet and zero-initialize the output projection layer $\mathbf{W}'_{O}$, ensuring that the model begins with no effect from these modifications, thus preserving its pre-existing behavior. Our design keeps the generative diffusion backbone untouched, effectively disentangling appearance control from motion generation. This separation allows the diffusion backbone to focus exclusively on pose control and dynamic synthesis, supported by ControlNet ~\cite{zhang2023adding} and temporal modules~\cite{guo2023animatediff}, while the dynamics adapter manages appearance consistency across frames.

\subsection{Implicit Local Face Expression Control}
\label{face}

In human video synthesis, natural variations in facial expressions significantly enhance realism and expressiveness. While many human image animation models offer robust control over full body poses, there has been limited efforts in simultaneously controlling facial expressions. Previous approaches to representing head motion often use simplified face landmark maps, capturing only key points such as the neck, nose, eyes, and ears. However, these simplified signals lack the detail needed for expressive facial animation. Moreover, even a basic facial skeleton encodes identity clues such as face shapes, which inadvertently influence face identity during cross-identity motion transfer. To address these limitations, we introduce \textbf{\xbody}, a control module in addition to body control, designed for identity-disentangled control over facial expressions and head poses, enabling more expressive and adaptable human video synthesis.

Inspired by X-Portrait~\cite{xie2024x},  instead of using an explicit face landmarks map from $\bm{I}_{i}$,  we crop the face patch and utilize a pre-trained portrait reenactment network $\mathcal{S}$  like FaceVid2Vid~\cite{wang2021facevid2vid} to transfer facial expressions onto a randomly selected subject with different facial attributes. This results in an identity-swapped face patch with close expressions, which is then reinserted at the original position of $\bm{I}_{i}$, with other pixels masked as blank, and used as the conditional input to an additional expression ControlNet $\mathcal{C}_F$ (Figure.~\ref{fig:pipeline}). Unlike explicit motion control signals, this cross-identity training approach enables $\mathcal{C}_F$ to learn identity-disentangled facial expressions and head movements implicitly from $I_{T}$, reducing appearance leakage from the driving signal. Notably, we bypass the need for $\mathcal{S}$ during inference, allowing expression control directly from the driving video.

\subsection{\mixdatatraining}
\label{fusion}
Prior human image animation models, especially those utilizing ReferenceNet for appearance control, generally mandate static backgrounds in training videos, which limits the capture of dynamic environmental details. On the other hand, collecting video data with both moving human and dynamic backgrounds for training is challenging.  We therefore introduce a mixed data training strategy, facilitating the diffusion backbone along with the temporal module to learn both human dynamics and background scene effects. Specifically, we integrate natural scene videos, such as waterfall, fireworks and wind, alongside real human motion videos for training.  For videos without human, we leave the conditional inputs to the Pose ControlNet $\mathcal{C}_{P}$ and \xbody $\mathcal{C}_{F}$ blank, enabling the model to generalize background motion independently. By using this mixed data, our model not only achieves more realistic dynamic details than those trained solely on human videos but also reduces unintended effects of ControlNets on background motion from the blank region of pose and expression conditional map.

\section{Experiments}
\label{exp}

\begin{table}[t!]\vspace{-3pt}
\centering
\caption{\textbf{Quantitative comparisons of \ourmodel with the recent state-of-the-art (SOTA) methods on dynamics texture generation}. A downward-pointing arrow indicates that lower values are better and vise versa. \textbf{DTFVD}~\cite{dorkenwald2021stochastic} is calculated by replacing the FVD pre-trained backbone with one trained on DTDB~\cite{hadji2018new}.  \textbf{FG-DTFVD} denotes the DTFVD is running on the \textbf{foreground} parts of the videos after segmentation, and \textbf{BG-DTFVD} denotes the DTFVD of the \textbf{background} parts.
}
\label{tab:dynamics}
\scalebox{0.79}
{\begin{tabular}{lcccc}
\toprule
Method & {\textbf{FG-DTFVD}} $\downarrow$  & {\textbf{BG-DTFVD}} $\downarrow$ & {\textbf{DTFVD}} $\downarrow$\\
\midrule
MagicAnimate~\cite{xu2024magicanimate} & \underline{1.753}

& 2.142&  2.601  \\
Animate-Anyone~\citep{hu2024animate} & 1.789 &
2.034 & \underline{2.310}  \\
MagicPose~\cite{chang2023magicpose} & 1.846 & \underline{1.901} & 2.412 \\
MimicMotion~\cite{zhang2024mimicmotion} &2.639 & 3.274 & 3.590  \\
\ourmodel &\textbf{0.900} & \textbf{1.101} &\textbf{1.518} \\
\bottomrule
\end{tabular}}
\vspace{-10pt}
\end{table}

\begin{table*}[t!]
\centering

\caption{\textbf{Quantitative comparisons of \ourmodel with the recent SOTA methods on human video animation}.  A downward-pointing arrow indicates that lower values are better and vise versa. \textbf{Face-Cos} represents the cosine similarity of the extracted feature by AdaFace~\cite{kim2022adaface} of face area between generation and ground truth image. \textbf{Face-Det} denotes the percentage rate of detected valid faces among all frames. $^*$ denotes the method is not open-sourced; hence, we used the unofficial implementation from~\cite{Moore-AnimateAnyone} to run their method for inference.
}
\label{tab:tiktok}
\scalebox{0.84}
{\begin{tabular}{lcccccccccc}
\toprule

\multirow{2}[2]{*}{Method}  & \multicolumn{8}{c}{\textbf{Foreground}} & \multicolumn{2}{c}{\textbf{Background}} \\ \cmidrule(lr){2-9} \cmidrule(lr){10-11}
   & \textbf{L1} $\downarrow$\  & \textbf{PSNR}\ $\uparrow$ & \textbf{LPIPS}\ $\downarrow$ & \textbf{SSIM}\ $\uparrow$ & \textbf{Face-Cos}\ $\uparrow$ & \textbf{Face-Det}\ $\uparrow$ & \textbf{FID}\  $\downarrow$ & \textbf{cd-FVD}\  $\downarrow$ & \textbf{FID}\ $\downarrow$ & \textbf{cd-FVD}\  $\downarrow$ \\
   
\midrule

MagicAnimate~\citep{chang2023magicpose}  &\underline{7.42e-05} &\underline{17.143} &\textbf{0.228} & \textbf{0.739} &0.297 &   \underline{92.1\%}&31.97  & 237.59 &38.86 & \textbf{176.17} \\
Animate-Anyone$^*$~\citep{hu2024animate} &11.8e-05 & 13.411 & 0.338 &  0.605& \underline{0.402} &   89.0\%& 33.75 & \underline{233.39} & 34.27& 203.59 \\
MagicPose~\citep{chang2023magicpose} &13.7e-05 &12.639 & 0.345 &  0.618 & 0.396&  85.5\%& \textbf{18.52}& 537.96  & \textbf{24.43}& 480.14 \\
MimicMotion~\citep{zhang2024mimicmotion}  &9.78e-05 &14.903 & 0.278&0.647 &0.193 &  92.0\%& 45.67& \textbf{150.01}  & 60.32& \underline{194.17} \\

\midrule

\ourmodel  &\textbf{7.15e-05}  & \textbf{17.201}  &\underline{0.249} &\underline{0.724} &\textbf{0.497} & \textbf{94.8\%} &\underline{22.56} &325.35  &\underline{25.59}  &281.78  \\
\bottomrule
\end{tabular}}
\vspace{-5pt}

\end{table*}

\subsection{Implementation Details}\label{implement}

\noindent \textbf{Dataset}\label{data} For animation of human videos, we train our model using a custom dataset including monocular camera recordings of 30-second human motions from 107,546 videos (900 hours in total) with both indoor and outdoor scenes. All the data were processed with a cropped resolution of 896$\times$512. Sequences of low quality were filtered out with ~\cite{hosu2020koniq}. All videos feature real subjects showcasing a diverse range of motions
and expressions in various scenes. For data processing, we follow the approach outlined in DisCo~\cite{wang2023disco,chang2023magicpose} but enlarge the cropping region to include the full body. For \mixdatatraining, we use Skyscape~\cite{Xiong_2018_CVPR} dataset, which contains 3000 time-lapse videos of dynamic sky scenes, e.g., cloudy skies and night scenes with moving stars. 

\noindent \textbf{Model Training and Inference}\label{model}
We utilize SD 1.5 as our generative backbone, and freeze its weights during the entire training phase.
Prior to training,  $\mathcal{C}_{P}$,  $\mathcal{C}_{F}$, and trainable parameters in $\mathcal{D}$ are initialized using SD 1.5, whereas the motion module is initialized with the weight of AnimateDiff~\cite{guo2023animatediff} . Our training is conducted in stages, where we first train $\mathcal{D}$, $\mathcal{C}_P$, and motion module with \mixdatatraining for five epochs. Then, we freeze these modules and train $\mathcal{C}_{F}$ for two epochs using human video data only.

An AdamW optimizer is utilized with a learning rate of $10^{-5}$ to train all modules. Each module undergoes training with 16 video frames in each step. During inference, we do not rely on the face-swapping network and directly feed the cropped local face patches from the driving video into \xbody.

\subsection{Evaluations and Comparisons}
\label{comp}
\noindent \textbf{Metrics.} We use three different groups of data for evaluation. 
1) To evaluate the overall human video generation quality, we use the test set split in TikTok~\cite{Jafarian_2021_CVPR_TikTok} proposed by DisCo~\cite{wang2023disco},  and report quantitative metrics \textbf{PSNR}, \textbf{SSIM}, \textbf{L1}, \textbf{LPIPS}, \textbf{FID} and \textbf{cd-FVD} for human foreground generation, and \textbf{FID} and \textbf{cd-FVD} for background generation. \textbf{cd-FVD} denotes \textbf{content-debiased-FVD}, a better Frechet Video Distance (FVD)~\cite{unterthiner2018towards} metric to reflect the overall generation quality, proposed by~\cite{ge2024content}. We also report Face Cosine Similarity (\textbf{Face-Cos}) to reflect the face identity preserving ability, following MagicPose~\cite{chang2023magicpose}. This metric is designed to gauge the model's capability to preserve the identity information of the reference image input. To compute this metric, we first align and crop the facial region in both the generated image and the ground truth. Subsequently, we calculate the cosine similarity between the extracted feature by AdaFace~\cite{kim2022adaface}, frame by frame of the same subject in the test set, and report the averaged value. These metrics has been widely used in previous  work~\cite{wang2023disco,xu2024magicanimate,chang2023magicpose,zhang2024mimicmotion}. In addition, we report the rate of detected faces among all frames in percentage, denoted as \textbf{Face-Det}.
2) To evaluate the dynamics detail generation quality, we use a self-collected test dataset from Pexels~\cite{pexels} with around 100 videos of 2 seconds each and report Dynamic Texture Frechet Video Distance (\textbf{DTFVD}) proposed by~\cite{dorkenwald2021stochastic}. DTFVD is calculated by replacing the pre-trained backbone network in FVD with one trained on Dynamics Texture Database (DTDB)~\cite{hadji2018new} for classification. This metric has also been widely used in other work~\cite{li2024generative,dorkenwald2021stochastic} to evaluate dynamics texture quality. We report DTFVD for both the whole videos and the background part of the videos after running human segmentation.
3) To further confirm the effectiveness of \ourmodel in dynamics detail generation, we conduct a comprehensive user study for comparison to other previous work~\cite{chang2023magicpose,zhang2024mimicmotion}. We collect 50 static real and synthetic reference images from Pexels~\cite{pexels} and generated by MidJourney~\cite{midjourney} to feed the model. We then ask users to judge the (1) dynamics quality of background nature, e.g., waterfall, fireworks, cloud, ocean, raining, snowing, grass, etc. (2) dynamics quality of human foreground, e.g., hair, clothes, etc. (3) appearance and identity preservation ability. 

\noindent \textbf{Quantitative Comparison}\label{quantitative}
We compare our method to the state-of-the-art diffusion model-based human video animation methods, including 1) CLIP embedding-based method DisCo~\cite{wang2023disco}; 2) ReferenceNet-based methods MagicPose~\cite{chang2023magicpose}, and MagicAnimate~\cite{xu2024magicanimate}; and 3) SVD-based method MimicMotion~\cite{zhang2024mimicmotion}. The main focus of this work is to improve the dynamics details generation quality.    
Tab.~\ref{tab:dynamics} presents a quantitative analysis of such quality. \ourmodel achieves significant improvements across different baseline models, indicating that the proposed method generates vivid expressiveness of dynamics.

Following previous work, we used sequences 335 to 340 from the TikTok~\cite{Jafarian_2021_CVPR_TikTok} dataset and additional self-collected videos by DisCo~\cite{wang2023disco} to test the animation ability of human subjects. Note that since the TikTok test set only contains indoor scenes, whose background is all static without any motions. 
Tab.~\ref{tab:tiktok} presents a quantitative analysis of human foreground subjects and background scenes from various methods, with segmentation using Segment Anything~\cite{kirillov2023segany}. The proposed methods achieve competitive performance across previous state-of-the-art methods, which indicates that the proposed method generates high-quality videos that align with human reference.

\begin{figure*}[t!]\vspace{-5pt}
\centering
 \includegraphics[width=0.99\linewidth]{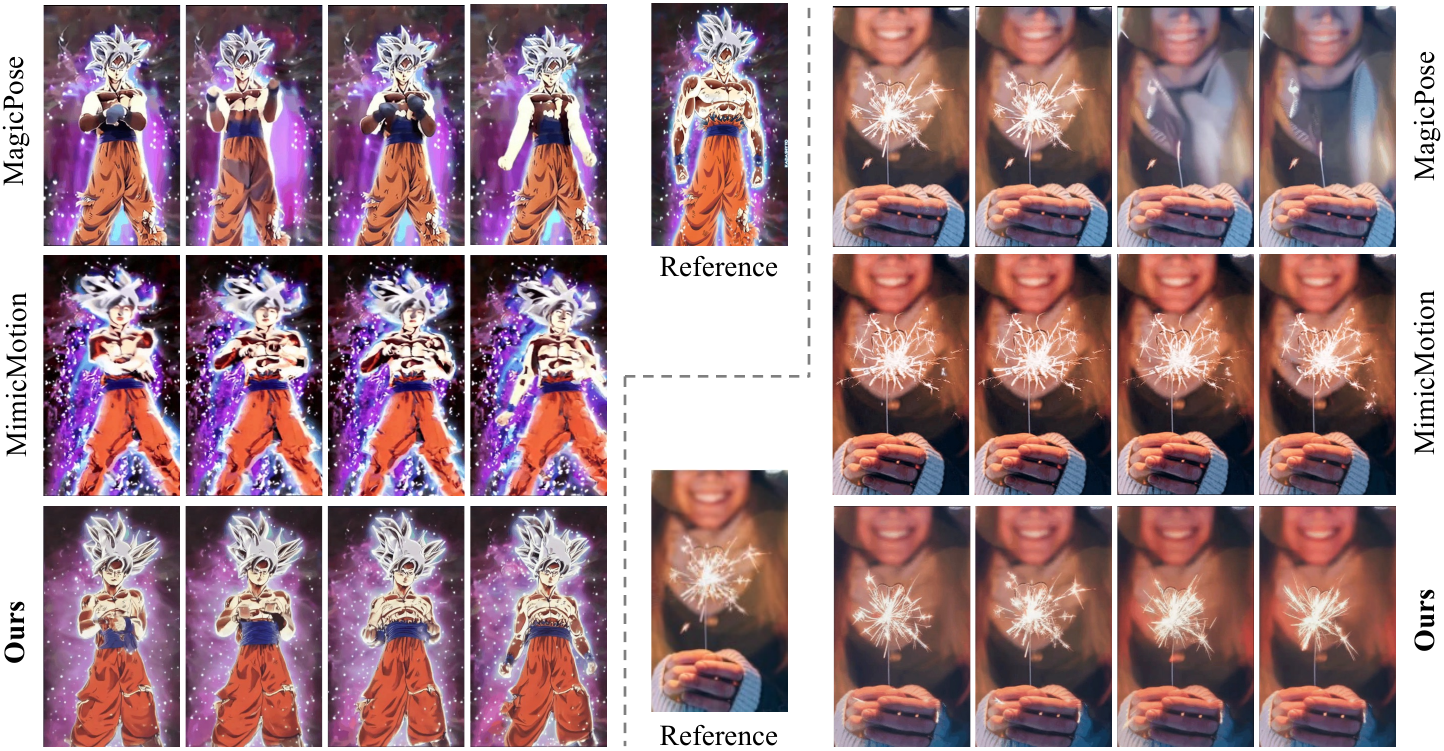}
   \caption{\textbf{Qualitative Comparison on Human in Dynamic Scene.} While existing SOTA methods struggle to generate consistent and realistic scene dynamics involving humans, our method successfully produces dynamic human-scene interactions while preserving the structure of the reference image.
   }
    \label{fig:comparisons}
\end{figure*}

\begin{figure*}[t!]
\vspace{-5pt}
\centering
 \includegraphics[width=0.98\linewidth]{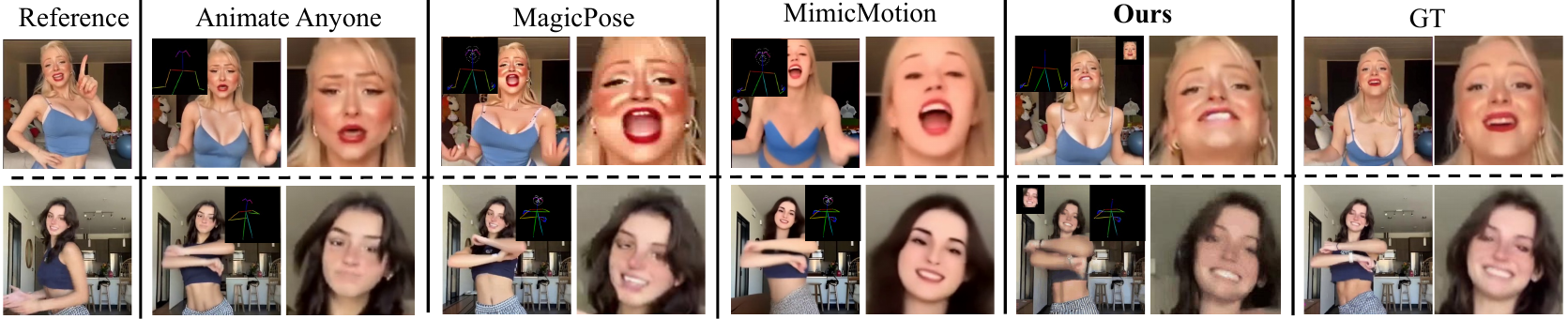}
   \caption{\textbf{Qualitative Comparison on Poses and Face Expressions Control.} We show each method on test cases using the same reference image and pose skeleton. For improved visualization, a zoomed-in view of the face area is also provided. Our method produces results that most closely match the ground truth and best preserve face identity.
   }
   \vspace{-5pt}
    \label{fig:tiktok}
\end{figure*}

\noindent \textbf{Qualitative Comparison}\label{qualitative}
We qualitatively compare the dynamics texture generation of \ourmodel with previous methods~\cite{chang2023magicpose,zhang2024mimicmotion} in Figure~\ref{fig:comparisons} and pose \& facial expressions control in Figure~\ref{fig:tiktok}. Note that MagicPose, MagicAnimate, and Animate-Anyone are recent representative works using \textbf{ReferenceNet}, and MimicMotion uses \textbf{SVD} for human video animation. Both MagicPose~\cite{chang2023magicpose} and MimicMotion~\cite{zhang2024mimicmotion} exhibit limited expressiveness, with most of their generated dynamics appearing almost static. Please refer to additional video examples provided in the supplementary materials for clearer observations and further comparison.

\noindent \textbf{User Study}\label{user}
 We provide a user study to compare \ourmodel with previous works~\cite{chang2023magicpose,zhang2024mimicmotion}. We collect reference images, pose conditions, and animation results from previous works and \ourmodel of 50 subjects as mentioned in Sec.~\ref{implement}. For each subject, we visualize different human poses and facial expressions and ask 100 users to rate the methods (from 0-5) according to the following three criteria: (1) dynamics quality of background nature (BG-Dyn) (2) dynamics quality of human foreground (FG-Dyn) (3) appearance and identity preservation ability (ID). We present the result of the average vote in Table~\ref{tab:user_study}. We observe that the user prefers \ourmodel more than ReferenceNet and SVD based works~\cite{chang2023magicpose,zhang2024mimicmotion}, especially in terms of dynamic texture generation. More details can be found in the supplementary material.

\begin{table}[t!]\vspace{-3pt}
\centering
\caption{\textbf{User study of \ourmodel}. We collect the ratings (0-5) from 100 participants for 50 test cases in the test set. We ask them to rate the generation in terms of Foreground Dynamics (\textbf{FG-Dyn}), Background Dynamics (\textbf{BG-Dyn}) and Identity Preserving (\textbf{ID}).
}
\label{tab:user_study}
\scalebox{0.9}
{\begin{tabular}{lcccc}
\toprule
Method & {\textbf{FG-Dyn}} & {\textbf{BG-Dyn}} & {\textbf{ID}} & {\textbf{Overall}}\\
\midrule
MagicAnimate~\cite{xu2024magicanimate} & \underline{2.34}

 & \underline{2.78}& 3.45 & 2.86 \\
Animate-Anyone~\cite{hu2024animate} & 2.21

 & 2.57 & \underline{3.89}  & \underline{2.89}\\
MagicPose~\cite{chang2023magicpose} & 2.23 & 2.18 & 3.85 & 2.75\\
MimicMotion~\cite{zhang2024mimicmotion} & 2.02 & 2.63 & 2.79 & 2.48 \\
\ourmodel  &\textbf{3.87} & \textbf{4.26} & \textbf{4.14}  & \textbf{4.09} \\
\bottomrule
\end{tabular}}
\vspace{-10pt}
\end{table}

\begin{table}[t!]
\centering
\caption{\textbf{Ablation Analysis of \ourmodel on dynamics texture generation and local facial expressions generation.} \textbf{w/RefNet} denotes \dynamicsadapter is replaced by a ReferenceNet. 
 \textbf{w/IP-A} denotes \dynamicsadapter is replaced by an IP-Adapter. 
 \textbf{w/lmk} denotes \xbody is not used for fine-tuning and face landmarks are used together with the pose skeleton.
 \textbf{wo/face} denotes \xbody is not used for fine-tuning. 
 \textbf{wo/fusion} denotes \mixdatatraining is not used for disentangled animation.
}
\label{tab:ablation}
\scalebox{0.8}
{\begin{tabular}{lccccc}
\toprule
Method & {\textbf{FG-DTFVD}}$\downarrow$ & {\textbf{BG-DTFVD}} $\downarrow$ & {\textbf{DTFVD}} $\downarrow$  & {\textbf{Face-Cos}} $\uparrow$\\
\midrule
 w/RefNet & 2.137
& 2.694& 2.823  & 0.466 \\
  w/IP-A & 3.738

 & 4.702& 4.851&  0.292 \\

w/lmk  & 0.914 & 1.125 & 1.589  &  0.406 \\
wo/face & \underline{0.912} & \textbf{1.098} & \underline{1.550}  & 0.442 \\
 wo/fusion & 1.301 & 1.467& 
1.652
 & \underline{0.495} \\
\ourmodel & \textbf{0.900}& \underline{1.101} &\textbf{1.518} &\textbf{0.497} \\
\bottomrule
\end{tabular}}
\vspace{-8pt}
\end{table}

\subsection{Ablation Analysis}\label{ablation}
In this section, a comprehensive ablation analysis of \ourmodel is presented. We evaluate the effectiveness of our face expressions and ID enhancement modules on TikTok~\cite{Jafarian_2021_CVPR_TikTok} test set in Tab.~\ref{tab:ablation}. 
To confirm the effectiveness of our \dynamicsadapter and \mixdatatraining, we quantitatively evaluate the DTFVD on our self-collected data and present the result in Tab.~\ref{tab:ablation}.

\subsection{Limitations and Future Works}
\label{limitation}
Despite its effectiveness in the dynamic expressiveness generation of human video animation, our \ourmodel has certain limitations, particularly in scenarios where the target pose significantly deviates from the reference human. For instance, during extreme zooming in or out, the appearance and identity may not be perfectly preserved. Additionally, our method struggles to generate perfect hand poses. We believe that these challenges can be addressed by collecting more high-quality data and employing advanced hand pose representations as input. 

In the future, we will explore applying \dynamicsadapter to more powerful base image and video diffusion models, such as SVD~\cite{blattmann2023stable}, SDXL~\cite{podell2023sdxl} and Stable Diffusion 3~\cite{esser2024scaling}, to achieve better performance. Moreover, we will investigate adding the camera trajectory or drag control proposed in~\cite{yin2023dragnuwa,wu2025draganything,he2024cameractrl,kuang2024collaborative} to our model so that we have a more user-friendly condition.

\section{Conclusion}
\label{conclusion}
In this work, we propose \textit{\ourmodel}, a photorealistic human video animation pipeline with the ability of consistent motion control and vivid dynamics details generation.  We propose an efficient \textit{\dynamicsadapter} module to preserve the human appearance reference while maintaining the foundation model's ability to generate high-quality dynamics. To boost the dynamics modeling capability further, we propose a \textit{\mixdatatraining} strategy, mixing the training data from real-human and natural scene videos.
Moreover, we incorporate two plug-in modules, an \textit{\xbody} for facial expressions editing and a \textit{\faceencoder} for face local identity preservation enhancement. Finally, all proposed modules can be treated as extensions to SD and used for customized pre-trained weights of SD-UNet. Extensive evaluation of various models also validates the effectiveness and generalizability of our model.

\noindent\textbf{Ethics Statement.} Our work aims to improve human image animation from a technical perspective and is not intended for malicious use like fake videos. Therefore, synthesized videos should clearly indicate their artificial nature.


{
    \small
    \bibliographystyle{ieeenat_fullname}
    \bibliography{main}
}
\clearpage
\newpage
\clearpage
\setcounter{page}{1}
\maketitlesupplementary

\section{Video Results}

We provide additional video results generated from \ourmodel in our project page.

\textbf{Comparison of Different Appearance Reference Module Designs:}
To demonstrate the effectiveness of our proposed \dynamicsadapter, we provide visual comparisons with IP-Adapter and ReferenceNet. Please refer to the \textbf{Different Architecture Designs} section for details.

 \textbf{Comparison to Previous Works:}
To evaluate the performance of \ourmodel in generating dynamic textures for human image animation, we present visual comparisons with previous state-of-the-art methods, including the ReferenceNet-based approach from~\cite{chang2023magicpose} and the SVD-based method from~\cite{zhang2024mimicmotion}. Details can be found in the \textbf{Comparison to Previous Works} section.

 \textbf{Ablation Study:}
To highlight the contribution of \mixdatatraining to our pipeline, we present a visualized ablation study. Please refer to the \textbf{Effectiveness of Mix data training} section of the project page.

\section{Quantitative Evaluation of Cross-Driving Reenactment}\label{cross_driven}
In this section, we present  quantitative evaluations for cross-driving video generation. We generated 200 videos for \ourmodel and each baseline method using various in-the-wild driving motions and reference images. The overall quality of cross-driving generation is assessed using DTFVD and FID metrics, comparing the distribution of the generated videos with the training videos. To evaluate the control accuracy of facial expressions, we crop the face area of both generated and driving videos and calculate their mean difference of face landmarks by MediaPipe~\cite{lugaresi2019mediapipe}.
The numerical results are summarized in Tab.~\ref{tab:cross_driving}, where \ourmodel demonstrates superior face expression control accuracy (Face-Exp) and dynamics (DTFVD), and comparable perceptual quality (FID).

\begin{table}[t!]\vspace{-3pt}
\centering
\caption{\textbf{Quantitative comparisons of \ourmodel with recent state-of-the-art (SOTA) methods on cross-driving human animation}. A downward-pointing arrow indicates that lower values are better. \textbf{DTFVD} and \textbf{FID} are used to evaluate the overall quality of generated videos. \textbf{Face-Exp} denotes the absolute error of facial expressions between generated videos and driving videos.
}
\label{tab:cross_driving}
\scalebox{0.79}
{\begin{tabular}{lcccc}
\toprule
Method & {\textbf{DTFVD}} $\downarrow$  & {\textbf{FID}} $\downarrow$ & {\textbf{Face-Exp}} $\downarrow$\\
\midrule
MagicAnimate~\cite{xu2024magicanimate} & 6.708 & 250.75 & 0.134  \\
Animate-Anyone~\citep{hu2024animate} & \underline{6.820} & 253.29 & 0.123\\
MagicPose~\cite{chang2023magicpose} & 7.062 & \textbf{244.25} & 0.121\\
MimicMotion~\cite{zhang2024mimicmotion} & 6.823 & 258.91 & \underline{0.109} \\
\ourmodel & \textbf{5.923} & \underline{246.16} & \textbf{0.105} \\
\bottomrule
\end{tabular}}
\vspace{-10pt}
\end{table}

\section{Details of User Study}\label{user_details}

In this section, we provide a comprehensive user study for qualitative comparison between \ourmodel and previous works~\cite{hu2024animate,xu2024magicanimate,chang2023magicpose,zhang2024mimicmotion}. 
We generate 50 different human animation results from all baseline models and \ourmodel, where the results are anonymized and shuffled.
On the online platform Prolific , we ask 100 users to rate these methods from 0(worst) - 5(best).


\textbf{Criteria for Judgment:}
Since our paper focuses on the dynamics of texture generation and motion control with human reference, the criteria for evaluation are (1) dynamics quality of background nature (BG-Dyn), (2) dynamics quality of human foreground (FG-Dyn), (3) appearance and identity preservation ability (ID). 

\textbf{Results and Statistical Analysis:}
The result is presented in Tab. 3 of the main paper. In addition, we perform a one-way analysis of variance (ANOVA) test on the ratings. ANOVA tests whether the means of multiple groups of data (methods in this case) are significantly different. For each metric, we compare the ratings across all five methods. Specifically,  \textbf{F-statistic} measures the ratio of variance between group averaged values to the variance within groups. A higher F-statistic indicates greater variability between group-averaged values relative to within-group variability.  \textbf{P-value}  tests the null hypothesis that all group means are equal. A small p-value (typically $\leq$ 0.05) indicates significant differences between groups. As reported in Tab.~\ref{tab:anova_results}, all metrics (FG-Dyn, BG-Dyn, ID, Overall) have p-values $\leq$ 0.05, indicating statistically significant differences between methods. The F-statistic for each metric shows the relative strength of these differences. \ourmodel consistently achieves the highest averaged ratings across all metrics (as seen in Tab. 3 of the main paper), and the differences are statistically significant. 


\begin{table}[h!]
\centering
\vspace{-5pt}
\begin{tabular}{lcc}
\hline
Metric & \textbf{F-statistic} &\textbf{ p-value} \\
\hline
FG-Dyn & 7.495 & 0.000007 \\
BG-Dyn & 5.327 & 0.000331 \\
ID     & 4.685 & 0.001016 \\
Overall & 5.617 & 0.000199 \\
\hline
\end{tabular}
\caption{ANOVA Test Results for Ratings from the User Study.}
\label{tab:anova_results}
\end{table}

\section{More Details on Prior Appearance Reference Control Designs}\label{more_pre}

\noindent\textbf{ReferenceNet} was initially introduced by Animate-Anyone~\cite{hu2024animate}. It adopts the same architecture as the Appearance Encoder in MagicAnimate~\cite{xu2024magicanimate} and the Appearance Control Model in MagicPose~\cite{chang2023magicpose}. Building upon prior advancements in dense reference image conditioning, such as the manipulation of self-attention layers in the UNet demonstrated by MasaCtrl~\cite{cao2023masactrl} and Reference-only ControlNet~\cite{Zhang}, ReferenceNet enhances identity and background preservation, significantly improving single-frame fidelity.
The naive self-attention calculation in the transformer blocks of the diffusion UNet can be represented as:
\begin{equation}
\label{eq:sd_naive_self_attn}
\bm{A}_i = \texttt{softmax}(\frac{\bm{Q}_{i} \bm{K}_{i}^\top}{\sqrt{d}}) \bm{V}_{i} , 
\end{equation}
However, ReferenceNet introduces a trainable duplicate of the base UNet, which computes conditional features from the reference image $I_{R}$ for each frame $I_{i}$. Unlike ControlNet, which integrates conditions additively in a residual manner, ReferenceNet injects the features derived from $I_{R}$ directly into the spatial self-attention layers of the UNet blocks. This is achieved by concatenating the reference features with the original UNet's self-attention hidden states. 
The process can be expressed as:
\begin{equation}
\begin{aligned}
\bm{A}_i = \texttt{softmax}(\frac{\bm{Q}_{i} \bm{K}_{i}^{'\top}}{\sqrt{d}}) \bm{V}_{i}^{'} , 
\end{aligned}
\end{equation} 

\begin{equation}
\begin{aligned}
    \bm{Q}_{i}&=W^{\bm{Q}_{i}}{z}_{{i}}, \bm{K'}_{i} = W^{\bm{K}_{i}}[{z}_{{i}},{z}_{r}], \bm{V'}_{i} = W^{\bm{V}_{i}}[{z}_{{i}},{z}_{r}],
\end{aligned}
\end{equation} 

where \([\cdot]\) denotes concatenation operation and ${z}_{i}, {z}_{r}$ denotes the self-attention hidden states from $I_{i}, I_{R}$. This self-attention mechanism strictly queries and preserves the information from the reference image in the denoising process, including human identity and background. 

\noindent\textbf{IP-Adapter}~\cite{ye2023ip} is composed of two key components: an image encoder that extracts features from the image prompt and adapted modules with decoupled cross-attention to integrate these features into the LDM UNet. A pretrained CLIP image encoder is employed to extract features from the reference image $I_{R}$.

To effectively decompose the extracted global image embedding, a lightweight trainable projection network—comprising a linear layer and Layer Normalization is utilized. This network projects the global image embedding into a sequence of features, ensuring that the dimensionality of the projected image features matches the dimensionality of the text features used in the UNet.

The integration of image features into the UNet is performed through adapted modules with decoupled cross-attention. In the original LDM, text features from the CLIP text encoder are incorporated into the UNet via cross-attention layers. In this setup, given the query features ${z_{r}}$ derived from $I_{R}$, the hidden states of the UNet for each frame $I_{i}$, and the text features ${z}_{t}$, the output of the cross-attention mechanism is defined as:

\begin{equation}
\begin{aligned}
\bm{A'}_i = \texttt{softmax}(\frac{\bm{Q}_{i}^{'} \bm{K}_{i}^{'\top}}{\sqrt{d}}) \bm{V}_{i}^{'} , 
\end{aligned}
\end{equation}

\begin{equation}
\begin{aligned}
    \bm{Q}_{i}^{'} = W^{\bm{Q}_{i}^{'}}{z}_{i}, \bm{K}_{i}^{'} = W^{\bm{K}_{i}^{'}}{z}_{t}, \bm{V'}_{i} = W^{\bm{V}_{i}^{'}}{z}_{t},
\end{aligned}
\end{equation} 

Then, another cross-attention layer for each original layer in the UNet is added to inject image features. Given the image features ${z_r}$, the output of this cross-attention is computed as follows:
\begin{equation}
\begin{aligned}
\bm{A''}_i = \texttt{softmax}(\frac{\bm{Q}_{i}^{'} \bm{K}_{R}^{'\top}}{\sqrt{d}}) \bm{V}_{R}^{'} , 
\end{aligned}
\end{equation}

\begin{equation}
\begin{aligned}
    \bm{Q}_{i}^{'} = W^{\bm{Q}_{i}^{'}}{z}_{i}, \bm{K}_{R}^{'} = W^{\bm{K}_{R}^{'}}{z}_{r}, \bm{V'}_{R} = W^{\bm{V}_{R}^{'}}{z}_{r},
\end{aligned}
\end{equation}

The same query $\bm{Q}{i}^{'}$ is shared between the image cross-attention and the text cross-attention mechanisms. As a result, only two additional trainable parameters, $W^{\bm{K}{R}^{'}}$ and $W^{\bm{V}_{R}^{'}}$, are introduced as linear layers for each cross-attention module.
The output of the image cross-attention is then combined with the output of the text cross-attention through a simple addition operation. Accordingly, the final formulation of the decoupled cross-attention is denoted as:
\begin{equation}
\begin{aligned}
\bm{Out'}_i =  \bm{A'}_i \ +  \ \lambda\bm{A''}_i ,
\end{aligned}
\end{equation}
where $\lambda$ is an adjustable parameter. When $\lambda = 0$, the model is the same as a frozen pre-trained LDM.

\noindent\textbf{Stable Video Diffusion (SVD)}~\cite{blattmann2023stable} is a diffusion-based video generation model that extends the latent diffusion framework originally designed for 2D image synthesis to produce high-resolution, temporally consistent videos from text and image inputs. SVD UNet introduces two types of temporal layers: 3D convolution layers and temporal attention layers, and temporal layers are also incorporated into the VAE decoder.
For training, the DDPM~\cite{DDPM} noise scheduler used in Stable Diffusion~\cite{rombach2022high} is replaced by the EDM~\cite{EDM} scheduler, alongside EDM's sampling method. Unlike traditional DDPM models that rely on discrete timesteps \(t\) for denoising, EDM uses a continuous noise scale 
\(\sigma_t\) By incorporating \(\sigma_t\) as input to the model, EDM enables more flexible and effective sampling, utilizing continuous noise strengths instead of discrete timesteps during the denoising process. This end-to-end training paradigm enhances temporal consistency in video generation.
However, SVD faces challenges when dealing with cross-driving cases. The reference image is concatenated with the noisy latent and directly input to the UNet, leading the model to deform the reference image into the first frame of the video rather than encoding the reference image and learning its semantic information implicitly, as achieved by ReferenceNet~\cite{hu2024animate}, IP-Adapter~\cite{ye2023ip}, and \dynamicsadapter. While fine-tuning the UNet, as in MimicMotion~\cite{zhang2024mimicmotion}, is a potential solution, it struggles to generalize to out-of-domain identities beyond the training data, as shown in Fig. 5 of our main paper and the supplementary videos.


\end{document}